\documentclass[10pt,letterpaper]{article}
\usepackage{cogsci}
\cogscifinalcopy
\usepackage[utf8]{inputenc} 
\usepackage[T1]{fontenc}    
\usepackage{microtype}
\usepackage{times,latexsym}
\usepackage{graphicx} \graphicspath{{figures/}}
\usepackage{amsmath,nicefrac}
\usepackage{acronym}
\usepackage[breaklinks,colorlinks,citecolor=black,bookmarks=true]{hyperref}
\usepackage{balance}
\usepackage{xspace}
\usepackage{setspace}
\usepackage[skip=3pt,font=small]{subcaption}
\usepackage[skip=3pt,font=small]{caption}
\usepackage[dvipsnames,svgnames,x11names,table]{xcolor}
\usepackage[capitalise,noabbrev,nameinlink]{cleveref}
\usepackage{booktabs,tabularx,colortbl,multirow,multicol,array,makecell,tabularray}
\usepackage{enumitem}
\usepackage[misc]{ifsym}
\usepackage{CJKutf8} 

\makeatletter
\DeclareRobustCommand\onedot{\futurelet\@let@token\@onedot}
\def\@onedot{\ifx\@let@token.\else.\null\fi\xspace}

\def\vs{\emph{vs}\onedot}

\makeatother

\makeatletter
\renewcommand{\paragraph}{%
    \@startsection{paragraph}{4}%
    {\z@}{0ex \@plus 0ex \@minus 0ex}{-1em}%
    {\normalfont\normalsize\bfseries}%
}
\makeatother

\usepackage[backend=biber,style=apa,natbib=true,annotation=false,backref=true]{biblatex}
\crefname{algorithm}{Alg.}{Algs.}
\Crefname{algocf}{Algorithm}{Algorithms}
\crefname{section}{Sec.}{Secs.}
\Crefname{section}{Section}{Sections}
\crefname{table}{Tab.}{Tabs.}
\Crefname{table}{Table}{Tables}
\crefname{figure}{Fig.}{Figs.}
\Crefname{figure}{Figure}{Figures}
\crefname{equation}{Eq.}{Eqs.}
\Crefname{equation}{Equation}{Equations}
\crefname{appendix}{Appx.}{Appxs.}
\Crefname{appendix}{Appendix}{Appendices}

\AtBeginDocument{
    \acrodef{ai}[AI]{Artificial Intelligence}
    \acrodef{rsa}[RSA]{Rational Speech Act}
    \acrodef{mrr}[MRR]{Mean Reciprocal Rank}
    \acrodef{mlp}[MLP]{MultiLayer Perceptron}
    \acrodef{coca}[COCA]{Corpus of Contemporary American English}
    \acrodef{coha}[COHA]{Corpus of Historical American English}
    \acrodef{acck}[Acc@k]{top-k accuracy}
    \acrodef{s1}[$S_1$]{Pragmatic Speaker model}
    \acrodef{s0}[$S_0$]{Literal Speaker model}
    \acrodef{l0}[$L_0$]{Literal Listener model}
}

\title{Rational Communication Shapes Morphological Composition}

\author{%
    Fengyuan Yang$^{1,2,3,4,5}$, Yongqian Peng$^{1,2,3,4,5}$, Yuxi Ma$^{1,2,4,5}$, Chenheng Xu$^{1,2,4,5}$, and Yixin Zhu$^{2,1,4,5\,\textrm{\Letter}}$
    \vspace{6pt}\\\normalfont
    \small $^1$ Institute for Artificial Intelligence, Peking University\quad{}
    \small $^2$ School of Psychological and Cognitive Sciences, Peking University\\
    \small $^3$ Yuanpei College, Peking University\quad{}
    \small $^4$ State Key Laboratory of General Artificial Intelligence, Peking University\\
    \small $^5$ Beijing Key Laboratory of Behavior and Mental Health, Peking University
}

\begin{document}

\maketitle

\begin{abstract}
Human languages expand vocabularies by combining existing morphemes rather than inventing arbitrary forms.
Communicative efficiency shapes lexical systems at multiple levels \citep{gibson2019efficiency}, yet morphological composition---combining morphemes through compounding or affixation---has rarely been modeled as a historically situated speaker choice among competing morpheme sequences, leaving unanswered why a language settles on one morpheme combination over other plausible alternatives.
We ask whether a trade-off between listener recoverability and speaker production cost can predict attested compositions over contemporaneously available alternatives.
Here we show, within the \ac{rsa} framework \citep{frank2012predicting,goodman2016pragmatic} using a time-indexed lexicon constructed from \ac{coha} and \ac{coca}, that across 4323 naturally occurring English compounds and derivations spanning 1820--2019, attested compositions are systematically ranked above unattested alternatives generated from contemporaneously available morphemes.
Models integrating semantic informativeness with production cost outperform semantic-only and cost-only baselines on \ac{mrr} and \ac{acck}, with the advantage of \ac{s1} over the semantic-only baseline growing as the candidate set expands, where meaning alone leaves morphological choice underdetermined.
These findings suggest that lexicalization reflects a communicative trade-off between expressiveness and efficiency, extending rational accounts of communication from utterance-level choice to the internal structure of words.

\textbf{Keywords:} morphology; composition; rational communication; computational modeling
\end{abstract}

\section{Introduction}

Human languages are shaped by the pressure to communicate efficiently: across syntax, semantics, and the lexicon, language structure reflects a systematic tendency to convey meaning while minimizing effort \citep{gibson2019efficiency,zipf1949human,salge2015zipf,jiang2024finding,jiang2025finding}. This pressure is visible at the lexical level, where word lengths correlate not with usage frequency alone but more precisely with contextual predictability \citep{piantadosi2011word}: words that are easier to anticipate tend to be shorter, as an efficient coding system would predict \citep{shannon1948mathematical,levy2008expectation,jaeger2010redundancy}. Word formation brings this pressure into sharp focus: when speakers need to name new concepts, they draw on existing morphemes, combining them via compounding or affixation to produce words whose meanings are recoverable from their parts \citep{algeo1980all,brinton2005lexicalization,xu2024word}.

\begin{figure}[t!]
    \centering
    \includegraphics[width=\linewidth]{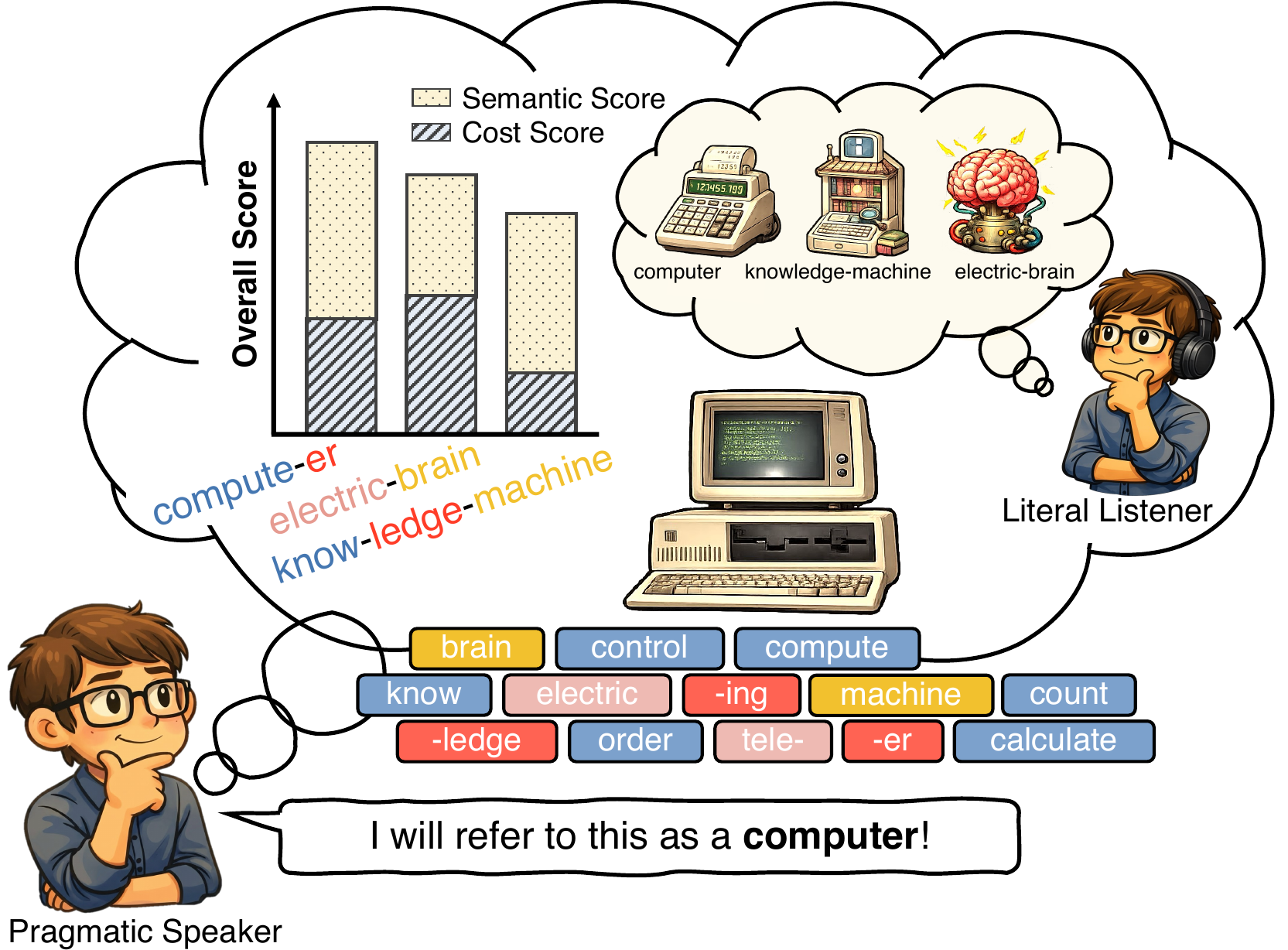}
    \caption{\textbf{Modeling morphological composition as rational communication.} A pragmatic speaker (left) constructs a word by selecting among candidate morpheme combinations available in the lexicon, balancing semantic informativeness---how well a literal listener (right) can infer the intended meaning from competing alternatives---against production cost. 
    The example shows alternative compositions for the concept “a programmable machine that performs arithmetic or logical operations,” including English \textit{compute-er}, Chinese \textit{electric-brain}, and Finnish \textit{knowledge-machine}.}
    \label{fig:teaser}
\end{figure}

We use \textit{morphological composition} to refer specifically to the combinatorial assembly of morphemes into new words, including compounds and derivations. This is a central subtype of \textit{word formation}, a broader domain that also includes back-formation, blending, and conversion \citep{vstekauer2005handbook,bauer2001morphological,plag1999morphological,peng2025probing}. Morphological composition is highly productive: speakers can coin new words by combining morphemes, and listeners can often interpret the results without prior exposure when component meanings are transparent. Experimental and computational studies confirm that semantic transparency systematically affects how novel compounds and derivations are interpreted \citep{reddy2011empirical,marelli2015affixation,mattiello2018morphosemantic,levin2019systematicity}. Yet strikingly different morpheme combinations across languages can name the same concept: English \textit{computer} uses \textit{comput-er}, while Chinese translates directly as \textit{electric-brain} and Finnish \textit{tietokone} as \textit{knowledge-machine} (see \cref{fig:teaser}). These divergences reflect not only what morphemes are available in a language's inventory but also what a listener can plausibly infer from them. Morphological composition is therefore not merely a structural property of language; it is a communicative choice made under efficiency pressures.

Prior computational and quantitative work already shows that morpheme selection is systematic rather than arbitrary. Morphological productivity research links the availability of word-formation patterns to lexical statistics, semantic transparency, and phonological or structural constraints \citep{bauer2001morphological,plag1999morphological,arndt2015word}. Analogical models capture competition among suffixation patterns such as \textit{-ity} \vs \textit{-ness} through gradient generalization over lexical exemplars \citep{arndt2015word}. Information-theoretic frameworks characterize global lexical systems as efficient encodings of semantic structure \citep{zaslavsky2018efficient,gibson2019efficiency}. Closest to our work, \citet{xu2024word} show that word reuse and combination support efficient communication of emerging concepts, but their model evaluates corpus-level lexical adoption rather than the speaker-side choice among alternative morpheme sequences for the same target meaning.

These frameworks illuminate important aspects of word formation, but leave open the candidate-level question we pursue here. Productivity and analogy models explain which word-formation patterns are available or preferred, not which particular morpheme sequence should be chosen when several could name the same concept. Information-theoretic accounts evaluate the efficiency of a lexical code as a whole rather than ranking specific morphological candidates available at a particular historical moment. \citet{xu2024word} ask which introduced forms survive in the lexicon, whereas we ask which morpheme combination is communicatively optimal as a historically situated choice. Addressing this requires a framework with an explicit candidate-comparison structure: one that ranks alternative morpheme sequences by how recoverable each is for a listener and how costly each is for a speaker.

The \ac{rsa} framework \citep{frank2012predicting,goodman2016pragmatic} provides exactly this. A \ac{s1} selects among explicit form alternatives by reasoning about listener recoverability and production cost, producing a ranked distribution over candidates via a utility $U$ that trades informativeness against cost, while a \ac{s0} supplies the semantic compatibility baseline from which pragmatic reasoning departs. This does not mean that \ac{rsa} should replace global efficiency, analogy, or adoption models; rather, it offers a complementary and interpretable decomposition for the moment-specific choice among candidate forms. \ac{rsa} has been applied to referential communication \citep{goodman2016pragmatic}, pro-drop \citep{chen2018modelling}, and cooperative explanation \citep{chandra2024cooperative}, yet morphology has remained largely untouched.

In this paper, we extend \acs{rsa}-style pragmatic modeling to morphological composition, treating word formation as a cooperative referential game. For a target concept $c$ at time $t$, a \ac{s1} selects among candidate morpheme sequences $u \in \mathcal{C}(c,t)$ by assigning utility $U(u,c)$ that trades listener recoverability against speaker production cost. Candidates are represented by per-morpheme-to-concept similarity statistics rather than a single composite embedding, preserving compositional structure while remaining order-agnostic. Using a time-indexed lexicon $\mathcal{L}_t$ built from WordNet \citep{miller1995wordnet} and historical \ac{coha}/\ac{coca} statistics, we evaluate 4323 English compounds and derivations spanning 1820--2019. Models that combine informativeness and cost outperform single-factor baselines, with the advantage growing as the candidate space expands.

\section{Computational Framework}

We treat existing poly-morphemic words as traces of historical lexical choices made around emergence time $t$. Empirically, $t$ is the first year with a non-zero \ac{coha}/\ac{coca} count; in the model, it approximates the lexical state near conventionalization. A time-indexed lexicon $\mathcal{L}_t$ supplies the shared linguistic background: word meanings, historical frequency, phonology, and form features. The central question is whether attested morpheme sequences rank above contemporaneous alternatives under a trade-off between recoverability and cost.

\paragraph{Notation.}
Let $c$ be a target concept, $u = (\mu_1, \ldots, \mu_m)$ a candidate morpheme sequence, and $\mathcal{C}(c,t)$ the candidate set available at time $t$. $S_0$ denotes semantic compatibility, \ac{l0} listener recoverability, \ac{s1} pragmatic speaker preference, and $\text{Cost}(u, \mathcal{L}_t)$ production cost. These are population-level operationalizations of \ac{rsa} terms, estimated from historical corpus resources rather than individual judgments. For a concept at lexical emergence, \ac{s1} conveys $c$ using morphemes from $\mathcal{L}_t$, balancing listener recoverability against production cost. We formalize listener interpretation, cost, and speaker choice in turn.

\subsection{Inferring listener interpretation}

To infer $L_0(c \mid u, \mathcal{L}_t)$, we define a literal production model $S_0(u \mid c, \mathcal{L}_t)$ capturing compatibility between utterance and concept. Following \citet{goodman2016pragmatic}, listener interpretation is:
\begin{equation}
    L_0(c \mid u, \mathcal{L}_t) \propto S_0(u \mid c, \mathcal{L}_t)\, P(c \mid \mathcal{L}_t),
\end{equation}
where $P(c \mid \mathcal{L}_t)$ is a concept prior. Since our targets are novel concepts without dedicated lexical entries, we assume a uniform prior over candidate concepts.

Rather than explicitly normalizing this posterior, we learn a compatibility function $f_\theta(u, c)$ intended to be monotonic with listener recoverability. Concretely, we compute similarities between the target concept embedding and the embeddings of morphemes in $u$. Rather than representing a candidate with a single composite embedding, we summarize the distribution of similarities between each morpheme and the concept using five statistics: mean, maximum, standard deviation, entropy, and the similarity between the whole candidate and the target. These feed the learned semantic scorer $f_\theta$.

To keep the search space tractable while preserving semantic plausibility, a gate $g(u, c)$ restricts candidate morphemes to a neighborhood of the target gloss $d_c$. Let $\text{kNN}(d_c)$ denote the morphemes among the $k$ nearest semantic neighbors of $d_c$:
\begin{equation}
    g(u,c) =
    \begin{cases}
        0, & \text{if } \exists\, \mu \in u \text{ such that } \mu \notin \text{kNN}(d_c), \\
        1, & \text{otherwise.}
    \end{cases}
\end{equation}
Candidates with $g(u, c) = 0$ are excluded before ranking, giving the gated compatibility function:
\begin{equation}
    \tilde{f}_\theta(u, c) =
    \begin{cases}
        f_\theta(u, c), & \text{if } g(u,c) = 1, \\
        -\infty,        & \text{if } g(u,c) = 0.
    \end{cases}
\end{equation}

\subsection{Estimating production cost}

Production cost is estimated from corpus- and form-based features in $\mathcal{L}_t$, including morpheme frequency, phonological complexity, and length. A learned function $h_\phi$ maps each morpheme's feature vector to a scalar cost, and candidate cost is additive:
\begin{equation}
    \text{Cost}(u, \mathcal{L}_t) = \sum_{i=1}^{m} h_\phi\big(\text{feat}(\mu_i, \mathcal{L}_t)\big),
\end{equation}
where $\text{feat}(\mu_i, \mathcal{L}_t)$ extracts the time-indexed feature vector for morpheme $\mu_i$. This treats frequent, shorter, and phonologically simpler morphemes as less costly, consistent with established links between predictability, form complexity, and production effort \citep{levy2008expectation,jaeger2010redundancy}.

\subsection{Integrating informativeness and cost}

Unlike $S_0$, \ac{s1} selects utterances by explicitly trading listener recoverability against production cost. In standard \ac{rsa} form:
\begin{equation}
    S_1(u \mid c, \mathcal{L}_t) \propto \exp\big(\log L_0(c \mid u, \mathcal{L}_t) - \text{Cost}(u, \mathcal{L}_t)\big).
\end{equation}
We implement this trade-off via a utility $U_\theta(u, c)$ defined over two learned base scores: gated semantic compatibility $\tilde{f}_\theta(u, c)$, approximating listener informativeness, and production cost $\text{Cost}(u, \mathcal{L}_t)$. The linear model combines these as:
\begin{equation}
    U_\theta(u, c) = \alpha \cdot \tilde{f}_\theta(u, c) - \beta \cdot \text{Cost}(u, \mathcal{L}_t) + b,
\end{equation}
where $\alpha$, $\beta$, and $b$ are learnable. The nonlinear model uses the same two inputs but allows interactions between informativeness and cost via a small \ac{mlp}:
\begin{equation}
    U_\theta(u, c) = \mathrm{MLP}_\theta\big(\tilde{f}_\theta(u, c),\, \text{Cost}(u, \mathcal{L}_t)\big).
\end{equation}
We refer to these as the Linear \ac{s1} and Nonlinear \ac{s1} models. The linear version closely follows the standard \ac{rsa} utility; the nonlinear version permits richer interactions between the two factors while retaining the same two-factor decomposition.

Taken together, the framework casts morphological composition as candidate ranking under a communicative trade-off: attested forms should rank highly when evaluated for recoverability and efficiency relative to alternatives in $\mathcal{L}_t$.

\section{Materials and Methods}

\subsection{Dataset}

The dataset contains 4323 English compounds and derivations from WordNet \citep{miller1995wordnet}, spanning first appearances from 1820 to 2019. Each item has a gloss $d_c$, a gold morpheme sequence segmented with MorphSeg \citep{morphseg}, and an emergence time $t$ defined as the first year with non-zero \ac{coha}/\ac{coca} count. First attestation is an imperfect proxy because lexicalization may lag concept emergence \citep{brinton2005lexicalization}; phoneme and syllable counts come from the CMU Pronouncing Dictionary.\footnote{\url{http://www.speech.cs.cmu.edu/cgi-bin/cmudict}; version 0.7b.}

The morpheme lexicon combines MorphSeg morphemes with an affix inventory \citep{affixdict}, assigning definitions from WordNet or the affix dictionary. Morpheme availability is quantified using cumulative type and token frequencies by decade for \ac{coha} \citep{davies2012corpus} and by year for \ac{coca} \citep{davies2010corpus}.

Candidate sets $\mathcal{C}(c,t)$ combine morphemes from WordNet synset lemmas, relational neighbors, and nearest neighbors from time-indexed word2vec spaces \citep{mikolov2013efficient}. We train skip-gram word2vec models separately on \ac{coha} decade slices (1820--2010) and \ac{coca} yearly slices (1990--2019), so that each target year draws distributional neighbors from its corresponding historical embedding space. We enumerate up to three morphemes per candidate, capped by length and candidate count, and apply the semantic gate $g(u,c)$ to avoid exhaustive enumeration over the full inventory.

\subsection{Feature representations}

\paragraph{Semantic features.}
Target glosses and morphemes are embedded with Qwen3-Embedding-8B \citep{zhang2025qwen3}, using both surface forms and available definitions for morpheme representations. For each candidate, the distribution of morpheme--concept cosine similarities is summarized by five statistics: mean, maximum, standard deviation, entropy, and the similarity between the whole candidate and the target. These form the 5-dimensional input to $f_\theta$.

\paragraph{Cost features.}
Each morpheme receives eight time-indexed features: character length, cumulative type and token frequency, 30-year token frequency, standalone and 30-year standalone word frequency, phoneme count, and syllable count. Frequency-derived features are clipped at zero and log-transformed before model input. All features are indexed by the target word's emergence year $t$.

\subsection{Model designs}

We evaluate five ranking models grouped into four families: cost-only, semantic-only, discriminative, and two \ac{s1} variants. All models are trained to rank the gold morpheme sequence above alternatives for the same target concept, with batches padded and masked so that only valid candidates and morpheme positions contribute.

\paragraph{Cost model.}
A two-layer per-morpheme \ac{mlp} $h_\phi$ maps each time-indexed cost feature vector $\mathbf{x}(\mu_i, t)$ to a scalar production cost. The candidate score is the negated sum over morphemes, so higher scores correspond to lower estimated cost:
\[
    \text{score}_{\text{cost}}(u) = -\gamma \sum_{i=1}^{m} h_\phi\!\bigl(\mathbf{x}(\mu_i, t)\bigr),
\]
where $\gamma > 0$ is a learnable scale parameter.

\paragraph{Semantic model.}
A three-layer \ac{mlp} $f_\theta$ maps the normalized semantic statistic vector $\mathbf{s}(u, c)$ to a scalar compatibility score, approximating $S_0$ and serving as the single-factor informativeness baseline:
\[
    \text{score}_{\text{sem}}(u, c) = f_\theta\!\bigl(\mathrm{LN}(\mathbf{s}(u, c))\bigr) / \tau,
\]
where $\tau > 0$ is a learned temperature that calibrates ranking confidence.

\paragraph{Discriminative model.}
Semantic statistics and masked-mean-aggregated cost features (5- and 8-dimensional, respectively) are concatenated into a 13-dimensional vector and scored by a single three-layer \ac{mlp}. This model serves as a flexible upper bound: it accesses all available information simultaneously but does not impose an interpretable informativeness--cost decomposition.

\paragraph{\ac{s1} models.}
Rather than consuming raw features, the \ac{s1} models operate as second-stage models over independently trained, \textit{frozen} Cost and Semantic base scorers. Freezing keeps the two \ac{rsa} components identifiable and prevents the \ac{s1} model from collapsing into an unconstrained discriminative ranker. The frozen scalar outputs are batch-normalized to a common scale and combined by a learned function $\psi$:
\[
    \text{score}_{S_1}(u, c) = \psi\bigl(\text{score}_{\text{cost}}(u),\; \text{score}_{\text{sem}}(u, c)\bigr).
\]
A \textit{linear} variant directly mirrors the \ac{rsa} utility $U = \alpha \log L_0 - \beta\,\text{Cost}$, while a \textit{nonlinear} variant uses a small two-layer \ac{mlp} to capture interaction effects between the two factors.

\subsection{Training and evaluation}

All models are trained as rankers with a pairwise softplus loss that encourages the gold morpheme sequence to score above sampled negatives, selected via curriculum hard-negative mining that emphasizes semantically similar candidates and those overlapping with the gold sequence. Models are optimized with Adam, gradient clipping, and early stopping on validation performance.

Data are split by year-stratified holdout to ensure historical periods remain represented in each partition: 20\% of items within each year are held out for testing, and the remainder are divided into training and validation using one fold of a 5-fold rotation scheme, yielding 864 held-out test items. Training and validation \ac{mrr} are used for fitting diagnostics and model selection. Final test metrics are computed on the held-out partition by ranking the gold sequence among up to 1024 candidates sampled from $\mathcal{C}(c,t)$, reporting \ac{mrr} and \ac{acck}.

\begin{figure}[b!]
    \centering
    \includegraphics[width=\linewidth]{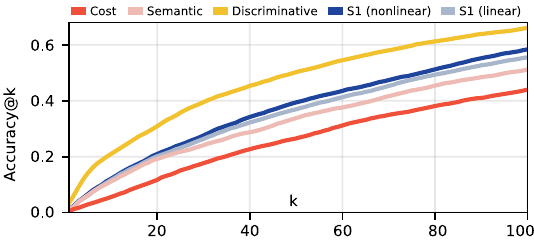}
    \caption{\textbf{Held-out \ac{acck} as a function of $k$.} The discriminative model consistently identifies the gold morpheme sequence most accurately. The advantage of \ac{s1} over the semantic-only baseline grows with $k$, consistent with pragmatic reasoning becoming most useful when many semantically plausible alternatives compete.}
    \label{fig:acc}
\end{figure}

\section{Results}

We first report held-out ranking performance, then use training and validation curves as diagnostics. Higher \ac{mrr} and \ac{acck} indicate that the attested morpheme sequence is assigned a better rank.

\subsection{Quantitative results}

\cref{fig:acc} summarizes held-out ranking performance across retrieval thresholds. The discriminative model remains strongest overall, but the theoretically important comparison is between the \ac{s1} models and the single-factor baselines. The \ac{s1} curves generally sit above both semantic-only and cost-only, with the clearest separation from semantic-only emerging at larger $k$: the regime in which many candidates are meaning-adjacent, semantic salience alone leaves the choice underdetermined, and production cost helps select among plausible forms.

The scalar \ac{mrr} results follow the same ordering. Cost alone is weakest (\ac{mrr}\,=\,0.031), semantic information is stronger (\ac{mrr}\,=\,0.047), and adding cost yields a modest further gain for the \ac{s1} models (\ac{mrr}\,=\,0.050--0.053). This gain is also visible at concrete retrieval thresholds: the nonlinear \ac{s1} model reaches Acc@10\,=\,12.08\% and Acc@20\,=\,20.94\%, compared with 11.10\% and 19.19\% for the semantic-only baseline. The discriminative model performs best overall (\ac{mrr}\,=\,0.096), indicating that additional feature interactions remain beyond the constrained two-scalar \ac{s1} decomposition.

\begin{figure}[t!]
    \centering
    \includegraphics[width=\linewidth]{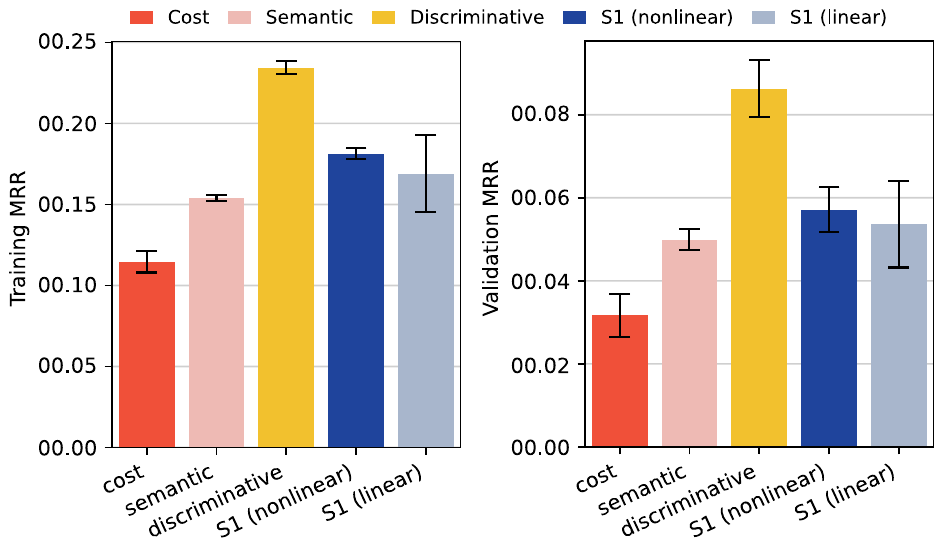}
    \caption{\textbf{Training and validation \acs{mrr} diagnostic.} The same broad ordering appears during fitting: models with access to semantic information outperform the cost-only baseline, and models that combine both sources of information perform best among the interpretable variants. Error bars indicate standard deviation across repeated training runs.}
    \label{fig:trainval}
\end{figure}

Training and validation behavior provides a diagnostic check on this held-out pattern. As shown in \cref{fig:trainval}, the same broad ordering appears during fitting: semantic information improves over cost alone, \ac{s1} improves over the single-factor baselines among the interpretable models, and the discriminative model remains strongest.

\begin{table*}[t!]
    \centering
    \caption{\textbf{Qualitative comparison of model rankings.} Gold morpheme sequences alongside the top-3 candidates produced by three models. The \ac{s1} model ranks the gold sequence at or near the top; the cost-only model selects high-frequency but semantically irrelevant fragments; the semantic-only model selects meaning-adjacent but morphologically odd forms.}
    \label{tab:qualitative}
    \setlength{\tabcolsep}{20pt}
    \resizebox{\linewidth}{!}{%
        \begin{tabular}{llll}
            \toprule
            \textbf{Model} & \textbf{Gold} & \textbf{Top-3 predictions} & \textbf{Rank} \\
            \midrule
            Linear \ac{s1}  & [`laundr', `y']     & [`laundr', `y'], [`laundr', `ify', `ing'], [`ca', `laundr', `launder']    & 1 \\
            Linear \ac{s1}  & [`affiliate', `ed'] & [`affiliate', `ation'], [`affiliate', `ed'], [`affiliate', `ation', `ed'] & 2 \\
            Linear \ac{s1}  & [`fiance', `ee']    & [`fiance', `eld'], [`fiance', `ee'], [`fiance', `ive']                    & 2 \\
            Semantic        & [`fiance', `ee']    & [`fiance', `est'], [`fiance', `eld'], [`person', `fiance']                & 5 \\
            Semantic        & [`cynic', `ism']    & [`cynic', `cynic'], [`cynic', `antipathy'], [`cynic', `ance']             & 83 \\
            Cost            & [`sauce', `pan']    & [`ed', `ing'], [`ed', `ing', `ing'], [`y', `ing', `ation']                & 74 \\
            Cost            & [`fiance', `ee']    & [`eld', `ly', `ity'], [`ed', `ly', `ency'], [`ly', `from', `er']          & 81 \\
            \bottomrule
        \end{tabular}
    }
\end{table*}

\subsection{Qualitative results}

\cref{tab:qualitative} illustrates the complementary failure modes of the single-factor baselines. The cost-only model favors short, high-frequency fragments---for instance, \texttt{ed+ing} for \textit{saucepan}---that are cheap but semantically uninformative. The semantic-only model errs in the opposite direction: for \textit{cynicism} it prefers the redundant \texttt{cynic+cynic}, and for \textit{fiancée} it selects semantically related but morphologically odd forms such as \texttt{person+fiance}.

The \ac{s1} model better balances these pressures, ranking \textit{laundry} first and placing the gold forms for \textit{affiliated} and \textit{fiancée} at rank 2, with near-misses that differ mainly in suffix choice. These examples show how integrating recoverability and cost favors candidates that are meaning-aligned without becoming unnecessarily long or morphologically implausible.

\section{Discussion}

The results support a conservative version of the rational-communication hypothesis. Semantic recoverability is the strongest single cue, but attested morphological forms rank better when recoverability is evaluated together with production cost. The effect is modest, as expected for a historically noisy word-formation problem with large candidate sets, but it is consistent across ranking metrics, qualitative examples, and temporal robustness checks. The discriminative model's advantage shows that lexicalization depends on additional structure beyond what the current \ac{rsa} operationalization captures; the \ac{s1} gains show that an informativeness--cost trade-off accounts for a systematic part of that choice.

\subsection{The cost-informativeness trade-off}

\begin{figure}[b!]
    \centering
    \includegraphics[width=\linewidth]{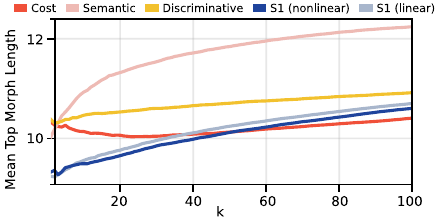}
    \caption{\textbf{Mean morphological length of top-$k$ predictions vs.\ $k$.} The semantic baseline consistently generates the longest candidates; \ac{s1} variants start shorter but grow steadily with $k$, eventually surpassing the cost-only model, reflecting a progressive relaxation of economy constraints to maintain semantic adequacy. Discriminative and cost-only models remain relatively stable.}
    \label{fig:length}
\end{figure}

The ablations suggest a clear division of labor. Cost alone captures economy but can select cheap, uninformative fragments; semantics captures recoverability but can overgenerate long or redundant forms (see \cref{tab:qualitative}). The \ac{s1} models combine these pressures, favoring candidates that remain meaning-aligned while avoiding costly or morphologically excessive alternatives. \Cref{fig:length} supports this interpretation. Semantic-only predictions are longest, consistent with maximizing semantic coverage without an economy term, while cost-only predictions remain short. The \ac{s1} curves begin closer to cost-only and lengthen as $k$ increases, suggesting that pragmatic ranking relaxes economy when extra morphological material helps preserve meaning.

The linear \ac{s1} weights point the same way. Averaged over checkpoints, both the semantic weight and the coefficient on the cost-based score are positive ($\alpha = 0.38$, $\beta_{\mathrm{score}} = 0.48$). Because the cost-based score is negated production cost, a positive $\beta_{\mathrm{score}}$ penalizes costly candidates. The ratio $\beta_{\mathrm{score}}/\alpha \approx 1.28$ depends on normalization and the learned base models, but confirms that both pressures remain active in the fitted \ac{rsa} utility.

This also explains why the \ac{s1} advantage is most visible beyond the top ranks. At low $k$, highly salient semantic cues dominate; as the candidate set expands, many alternatives become plausible, and cost helps select among them. The pragmatic model is therefore most useful precisely when morphological choice is underdetermined by meaning alone.

\subsection{Assumptions about speaker and listener knowledge}

Our framework estimates \ac{rsa} quantities from aggregate corpus statistics rather than individual judgments. Semantic compatibility is operationalized via distributional similarity, while production costs reflect population-level morpheme frequencies and form complexity. Thus $\mathcal{C}(c,t)$ should be read as a historically available community-level candidate space, not as a set of alternatives explicitly considered by any single speaker.

This is a system-level use of \ac{rsa}. Like information-theoretic efficiency accounts \citep{zaslavsky2018efficient,xu2024word}, it treats communicative utility as a pressure on the language system rather than a direct measurement of an individual. Its contribution is to make candidate comparison explicit: for a target concept at time $t$, which available morpheme sequence best balances listener recoverability and production cost? Future behavioral work can test whether speakers exhibit the same trade-off when coining novel morphological forms.

\subsection{Diachronic effects and the role of \texorpdfstring{$\mathcal{L}_t$}{}}

\begin{figure}[t!]
    \centering
    \includegraphics[width=\linewidth]{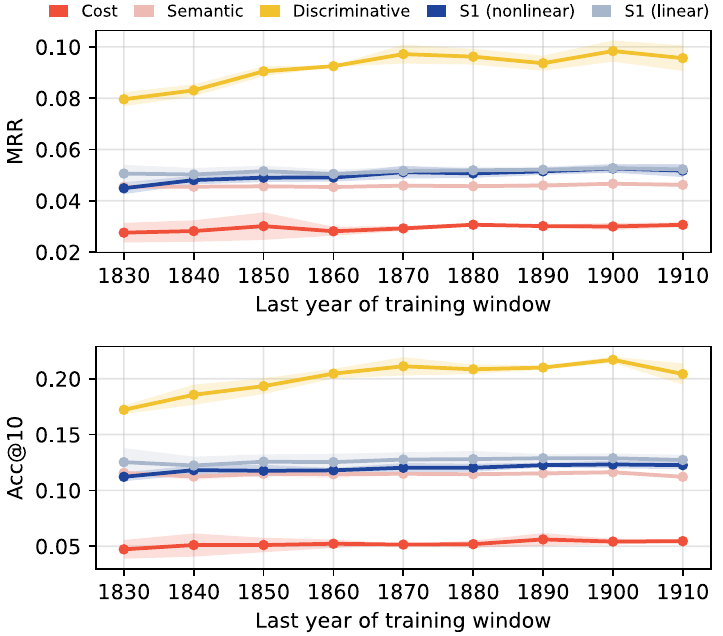}
    \caption{\textbf{Temporal robustness across cumulative training windows.} Ranking performance as a function of the last year included in the cumulative training window. \ac{mrr} (left) and Acc@10 (right) show stable model ordering across windows; shaded bands denote $\pm 1$ standard deviation across folds.}
    \label{fig:temporal}
\end{figure}

Conditioning candidate availability and production cost on $\mathcal{L}_t$ gives the model a diachronic mechanism: as morpheme inventories, frequencies, and distributional neighborhoods change, the most efficient composition for a given concept may shift accordingly. To test whether the main ordering holds across the historical range, we retrained all model families on cumulative windows ending between 1830 and 1910 and evaluated each on its matched held-out set (\cref{fig:temporal}). The broad ordering is stable: the discriminative model performs best, \ac{s1} remains above the single-factor baselines, and cost-only remains lowest. Because data coverage varies across windows and semantic scoring relies on contemporary embeddings, this should be read as a robustness check rather than a full model of diachronic semantic change.

The main temporal limitation is that semantic scoring still relies on contemporary Qwen3 embeddings, so semantic drift is represented only indirectly through candidate generation and historical frequency features. Diachronic semantic encoders or historical definition sources would allow a more faithful reconstruction of listener expectations at time $t$.

\subsection{The \texorpdfstring{\acs{rsa}}{RSA}--discriminative gap}

The discriminative model's advantage reveals where the current \ac{rsa} operationalization is too compressed. It has access to the full 13-dimensional feature vector and can learn interactions among semantic and cost features, whereas the \ac{s1} models combine only two frozen scalar scores. This design preserves the interpretable \ac{rsa} decomposition but discards cross-feature interactions, such as cases where a rare morpheme is acceptable because it is highly diagnostic for the target concept.

We therefore interpret the discriminative advantage as evidence that morphological composition depends on structure beyond a two-term utility. A natural next step is to enrich the \ac{rsa} model without abandoning its decomposition, for example by jointly fine-tuning the semantic and cost components, incorporating morphotactic constraints, or allowing morpheme cost to depend on contextual informativeness.

\section{Limitations and Future Directions}

\paragraph{Morphological structure.}
Our model treats candidate morpheme sequences as unordered sets, ignoring morphotactic constraints, affix ordering, and head-modifier structure. Incorporating ordering constraints or sequence-sensitive composition functions would better reflect speakers' knowledge of morphological well-formedness and reduce structurally implausible candidates.

\paragraph{Historical and temporal grounding.}
Semantic compatibility is estimated using Qwen3 embeddings trained on contemporary corpora, and concept emergence is proxied by first attestation rather than by earlier conceptual emergence, which lexicalization may lag. Both approximations introduce uncertainty into $\mathcal{L}_t$ reconstruction, compounded by uneven corpus coverage across decades. Models such as diachronic embedding \citep{ma2025word} would address these gaps.

\paragraph{Cross-linguistic generalization.}
We evaluate exclusively on English. Because compounding, affixation, and the informativeness--cost balance differ substantially across isolating, fusional, and agglutinative languages, extending the framework to typologically diverse languages is necessary to assess whether the communicative principles are universal.

\section{Conclusion}

We develop a computational account of morphological composition as rational communication, formalizing word formation as an \acs{rsa}-style choice among competing morpheme sequences in a time-indexed lexicon $\mathcal{L}_t$. The central prediction is that languages prefer compositions that are both semantically recoverable for a listener and efficient for a speaker, capturing the classic informativeness--cost trade-off.

Ranking experiments on 4323 naturally occurring English compounds and derivations support this prediction. Models integrating semantic compatibility with production cost outperform single-factor baselines on the year-stratified held-out split, while the discriminative model marks the remaining predictive signal outside the constrained \ac{s1} decomposition. Qualitative comparisons further show that pragmatic integration avoids the characteristic failure modes of single-factor approaches, prioritizing candidates that are meaning-aligned without becoming unnecessarily long or morphologically odd.

Taken together, our results suggest that morphological composition can be modeled as an emergent optimization shaped by the inventory and usage statistics of the historical lexicon, extending the rational communication program from utterance-level phenomena to the internal structure of words. Whether the same principles generalize across languages and word-formation types remains an open question.

\paragraph{Acknowledgement}

\cref{fig:teaser} was produced with AI-assisted illustration tools; all other figures were generated programmatically from experimental data. The authors would like to thank Guangyuan Jiang (MIT), Hongjie Li (PKU), Prof. Yanchao Bi (PKU), Prof. Huichao Yang (Hebei Normal University), and Dr. Qian Wang (PKU) for many valuable discussions and support. This work is supported in part by the National Natural Science Foundation of China (32595491, 62376009), the PKU-BingJi Joint Laboratory for Artificial Intelligence, the Wuhan Major Scientific and Technological Special Program (2025060902020304), the Hubei Embodied Intelligence Foundation Model Research and Development Program, and the National Comprehensive Experimental Base for Governance of Intelligent Society, Wuhan East Lake High-Tech Development Zone.

\printbibliography

\end{document}